\documentclass{llncs}
\usepackage{makeidx}  
\usepackage[font=small,labelfont=bf]{caption}
\usepackage{subcaption}
\usepackage{amsfonts}
\usepackage{amssymb}
\usepackage{amsthm,amsmath}
\usepackage{mathrsfs}
\usepackage{indentfirst}
\usepackage{multirow}
\usepackage{times}
\usepackage{epsfig}
\usepackage{graphicx}
\usepackage{caption}
 \usepackage{upgreek}
\usepackage[ruled,vlined]{algorithm2e}
\usepackage{algorithmicx}

\algnewcommand{\algorithmicforeach}{\textbf{for each}}
\algdef{SE}[FOR]{ForEach}{EndForEach}[1]
  {\algorithmicforeach\ #1\ \algorithmicdo}
  {\algorithmicend\ \algorithmicforeach}
 
\usepackage{subcaption}
\usepackage{graphicx}
\usepackage{textcomp}
\usepackage{soul}
\usepackage{algorithmicx}
\usepackage[justification=centering]{caption}
\usepackage{color,xcolor}
\usepackage{soul}

\begin{document}
\frontmatter          
\pagestyle{headings}  
\addtocmark{Real2Sim} 
\mainmatter              
\title{Real2Sim or Sim2Real: Robotics Visual Insertion using Deep Reinforcement Learning and Real2Sim Policy Adaptation}
\titlerunning{Real2Sim}  
%

\author{Yiwen Chen \and Xue Li *
Sheng Guo * \and Xian Yao Ng \and Marcelo H. Ang, Jr.
\thanks{*Xue Li and Sheng Guo are equally contributed and Sheng Guo is the corresponding author. E0576004@u.nus.edu}}
\authorrunning{Yiwen Chen et al.}
%
%
\tocauthor{Yiwen Chen, Xue Li, Sheng Guo, Xian Yao Ng, Marcelo H. Ang, Jr.}
\institute{Advanced Robotics Center, National University of Singapore, Singapore,\\ 
\email{yiwen.chen@u.nus.edu}
}


\maketitle              

\begin{abstract}
Reinforcement learning has shown a wide usage in robotics tasks, such as insertion and grasping. However, without a practical sim2real strategy, the policy trained in simulation could fail on the real task. There are also wide researches in the sim2real strategies, but most of those methods rely on heavy image rendering, domain randomization training, or tuning.
In this work, we solve the insertion task using a pure visual reinforcement learning solution with minimum infrastructure requirement. We also propose a novel sim2real strategy, Real2Sim, which provides a novel and easier solution in policy adaptation. We discuss the advantage of Real2Sim compared with Sim2Real.
\end{abstract}
\section{Introduction}
Insertion is an essential step in many assembly processes in manufacturing lines. Insertion task automatically executed by industrial robots is a frontier topic that has been studied for a long time. In this work, we are motivated to achieve comparably good performance but minimize the infrastructure cost (from more than \$500 to \$10), prerequisites, human labor intensity, and maximize the noise resistance.

\subsection{Minimum Infrastructure and Prerequisites}

Traditional control methods have high requirements for the precision of sensors and robots. The deployment of such methods usually depends on the complex and tuned setup. Some strict preconditions are assumed, such as the position of the hole being fixed and known \cite{10.1115/DSCC2015-9703}, target pose of the gripper is assumed given \cite{vecerik2018leveraging}, target image \cite{marven1} with the peg already in the hole, or the background color is simple \cite{8794074} and flatten \cite{schoettler2020metareinforcement}, or light quality is fixed \cite{marven1}. As for additional sensors, some methods use high-precision depth camera \cite{schoettler2020metareinforcement}, torque-force sensors \cite{8}
Those complex infrastructure and prerequisites severely limit practical applications. Noise and disturbance in our daily environments may easily defeat the fragile pre-assumptions. Therefore, there is an urgent need to design a method with minimum infrastructure and prerequisites. In our work, only one RGB low-cost camera is used to realize robotics insertion tasks instead, quite simple and adaptive.
%
%
Due to the expensive human-labor cost, we hope there is a solution that involves as little as possible human labor intensity. Generally, data labeling such as trajectory generation supervised training \cite{10} and detection training \cite{8794074}, or human demonstrations \cite{8794074} requires human intervention. Therefore, in this work, we also aim to minimize human intervention with unsupervised method.

\vspace{-0.5cm}
\begin{figure}
     \centering
     \begin{subfigure}[b]{0.6\linewidth}
         \centering
         \includegraphics[width=\textwidth]{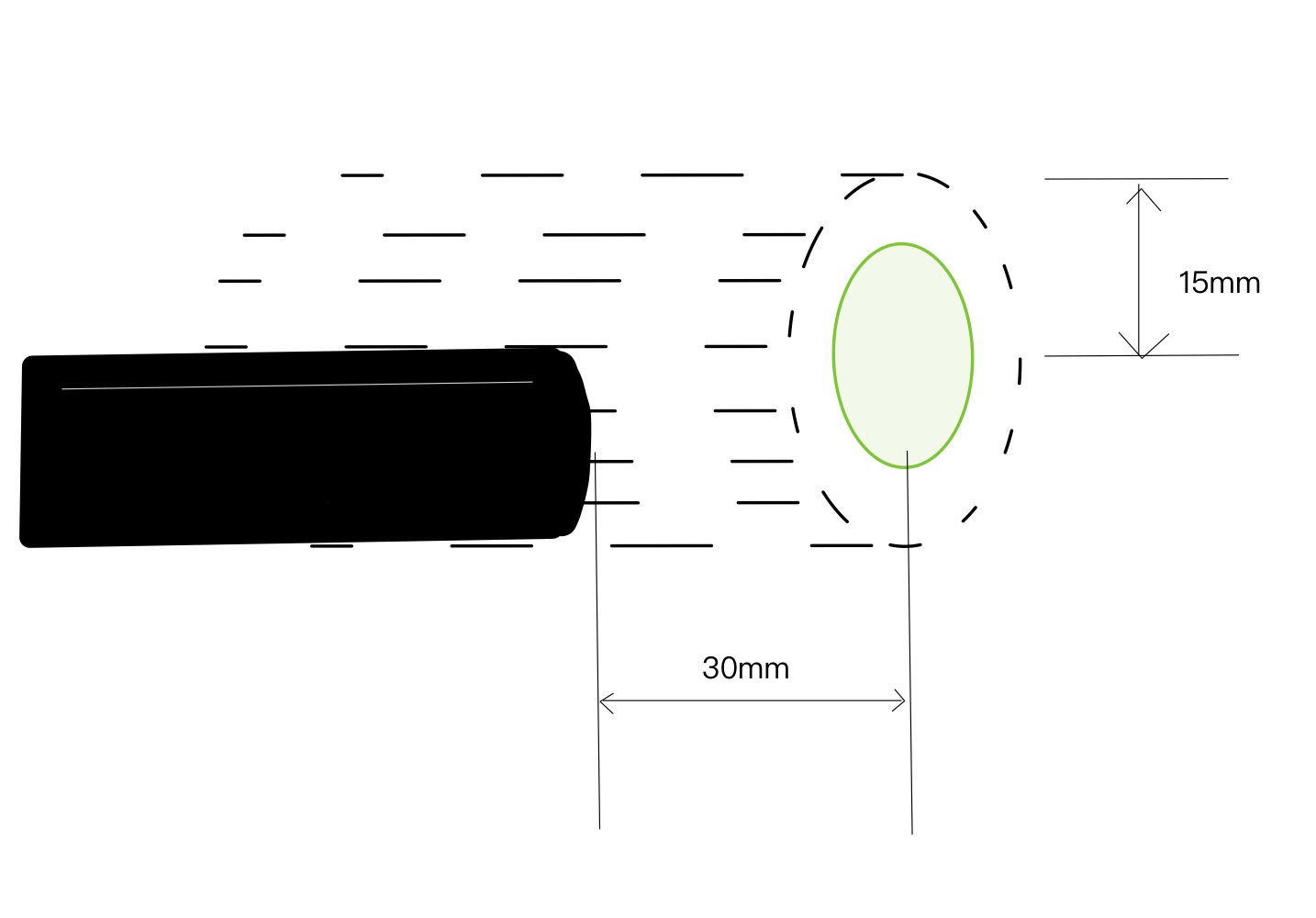}
         \caption{Insertion Start}
         \label{Task Statement}
     \end{subfigure}
     \hfill
     \begin{subfigure}[b]{0.35\linewidth}
         \centering
         \includegraphics[width=\textwidth]{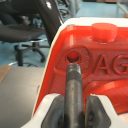}
         \caption{Insert with RGB}
         \label{Image Sample}
     \end{subfigure}
     
    \caption{Task Description}
    \label{Task Statement}
\end{figure}
\vspace{-0.5cm}

\subsection{Real2Sim or Sim2Real}

There has been increasing research work on robotic insertion focusing on learning-based methods \cite{marven1}\cite{camilo2020variable}. However, simulation learning-based methods may result in a serious decrease in the success rates when deployed to real situations without an effective sim2real strategy \cite{camilo2020variable}. Therefore sim2real adaptation efforts have been made to narrow the gap between the simulation and the reality\cite{rlcyclegan}\cite{tobin2017randomdomain}.

However, as for the sim2real strategy, the challenge here is that it is hard to exactly rendering the real-world industrial scene in the simulator. Previous sim2real work tries to enrich the high-quality simulation to match the real world in the training phase, which requires a huge effort in data preparation and costs significantly more. Domain randomization has also been discussed, but it may still fail when there is a huge gap between simulation and the real environment. With a generative model, it is also hard to control the quality of generated image\cite{rlcyclegan}. 

Our work is unique by applying an inverse idea, real2sim in the test phase. The real scene will be transformed into an image akin to the simulator before it is used as the input for insertion, as shown in Fig  \ref{fig:domain_adaptation_and_mix}.

\subsection{Contributions}

Summarily, our main contributions are as follows:
\begin{itemize}

\item With minimum infrastructure and prerequisites, we solve the robotics insertion problem with raw images from a monocular RGB camera. Our approach significantly reduced the infrastructure cost from more than \$500 to only \$10, and minimize the human intervention from heavy labeling to unsupervised training to achieve a comparable result with the benchmarks.

\item A unique real2sim approach is proposed to solve the last mile of the algorithm implementation and cancels the background and environmental noise. It provides a brand novel view in policy adaptation.

\end{itemize}

\section{Task Description}
The objective of this work is to develop a novel and practical approach applicable to the insertion task with uncertainty of hole pose, with only given a real-time camera video stream (RGB Images). The algorithm should be trained in virtual environment, the simulation, and tested in reality. 

A manipulator mounted with a real-time camera on the hand, capturing images in Fig \ref{Image Sample}, is used for this insertion task. Camera is the only observation sensor. The relative position and orientation between the camera and gripper is fixed and known. The diameter of the hole is 11mm, with 1mm clearance to the bolt.
The bolt is assumed to be firmly held by two fingers of the robot all the time. The initial position of the bolt in the gripper falls randomly in the range of a cylinder concentric with the hole, with a diameter of 20mm and a height of 3cm. During the whole process, the insertion speed in z direction (hole axis) keeps constant while the speed of x and y directions are provided by a policy network (deep neuron network). If the insertion depth reaches 5mm, the task is considered as successfully completed.
\begin{figure*}
    \centering
    \includegraphics[width=0.9\textwidth]{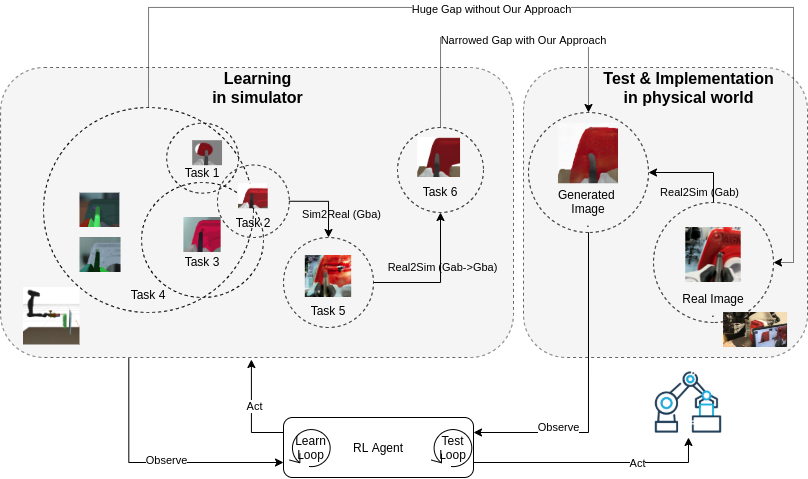}
    \caption{Insertion Task Domain Adaptation and Mixing Task Learning}
    \label{fig:domain_adaptation_and_mix}
\end{figure*}

\section{Related Works\label{literature_review}}
\subsection{Robotics Insertion}

Traditional methods of insertion tasks considerably rely on human ingenuity. The standard control method, such as the PID controller, heavily depends on engineers' manual modeling and tuning, which lacks both accuracy and adaptability while huge efforts and time are consumed \cite{marven1}. Active compliance control, though greatly contributes to the accuracy of the insertion task, remains unpractical to apply in most industrial scenes \cite{camilo2020variable}. By contrast, deep reinforcement learning enables robotics to learn insertion skills, solving insertion tasks efficiently and robustly \cite{8454796}.
And \cite{vecerik2018leveraging} proposed DDPGfd (Deep Deterministic Policy Gradient from Demonstration). However, strong prerequisites given in \cite{vecerik2018leveraging} limit the applicability in noisy reality. 
Different from specializing on one or a few specific insertion tasks, \cite{schoettler2020metareinforcement} propose an off-policy meta reinforcement learning method named probabilistic embeddings for actor-critic RL (PEARL), which enable robotics to learn from the latent context variables encoding salient information from different kinds of insertion, resulting in a rapid adaptation when doing a new task.    
In spite of the success of multi-stage insertion tasks, significant challenges are remained for further applying to a more complex environment.

\subsection{Reinforcement Learning and Sim2real Adaptation}
Model-based Reinforcement Learning and Model-free RL have drawn wide attention. Trust-region methods in reinforcement learning have been broadly used in solving robotics control and complex strategy learning \cite{ppopaper}\cite{trpo_paper}.
Domain randomization \cite{tobin2017randomdomain} shows that it is possible to transfer a simulated algorithm to a real-world scenario, which is also extensively discussed in the transfer learning research field. As for the combination of Sim2Real and deep RL, \cite{rlcyclegan} considered improving Cycle GAN with Q loss. Another notable sim2real domain adaptation research given in recent years is the cubic solving with RGB image and robotics hands\cite{cubic}. It leverages the advantages of deep reinforcement learning and domain adaptation methods called automatic domain randomization (ADR). GAN methods and domain randomization are also discussed in our approaches.

\section{Our Approach}

\subsection{Overview}

The main idea of our approach is to successfully learn the insertion action, called policy, in the simulation, and then to transfer this strategy, by building the sim2real and real2sim transferring pipeline with generative model and domain randomization tricks, to implement it into a real robotics insertion scenario. As for the learning part, we adopt mixing learning to increase its stability and robustness for various tasks and improve training efficiency. The algorithm is described in the algorithm \ref{algo}. In the inference part, the real physical robot would take it is as working under simulation, which is generated by the GAN model, even though it is working in a real physical world.

\begin{algorithm}
\SetAlgoLined
\SetKwBlock{Thread}{Thread}{end}
\SetKwBlock{Function}{Function}{end}

 Initialize$:$\;

 Trained GAN model $G_{\phi ab}$ and $G_{\phi ba}$\;
 Defined Tasks $T_1, T_2, T_3, T_4, T_5(G_{\phi ba}), T_6(G_{\phi ab},G_{\phi ba})$\;
 Vectored Mixing Task $VT= [T_1,T_2,T_i...]$\;
 Policy Model $a\sim\pi_\theta(a|o)$\; 
 Simulating observation $o_B$\;
 Real observation  $o_A$;

    \While{Training Loop}{
    Sample action $a\sim\pi_\theta(a|o_B)$\;
    $o$, $r$ = $VT$ ($a$)\;
     \If{reach training steps}{
      $\theta_{new} =$ Optimize PPO objective wrt $\theta$\;
      $\theta=\theta_{new}$\;
      }
    }
    \While{Inference Loop}{
        get $o_A$ from RGB camera\;
        Generate image $o_B= G_{\phi ab}(o_A)$\;
        Sample action $a\sim\pi_\theta(a|o_B$)\;
        Input $a$ to robot action server\;
    }

 \caption{\label{algo} Task Mixing Learning and Real2Sim}
\end{algorithm}

\subsection{Reinforcement Learning for Insertion Action Learning}
\subsubsection{Proximal Policy Optimization (PPO)}
In this approach, we adapt the advantage of PPO algorithm\cite{ppopaper}. The main idea of PPO is to restrict the policy update stepsize, by using a clipping factor \(\epsilon\). TRPO approach \cite{trpo_paper} gives objective function as following

\[r_t = \left[\frac{\pi_{\theta}\left(a_{t}|s_{t}\right)}{\pi_{\theta_{old}}\left(a_t|s_t\right)}  \right]\]

\[L_{(\theta)}=\hat{E_t} \left[ r_t(\theta)\hat{A}_t \right]\]

And PPO modified this objective function with clipping factor \(\epsilon\), making is much more easier to implement\cite{ppopaper}. The new objective function is defined as following. 
\[L_{(\theta)}=\hat{E_t} \left[min( r_t(\theta)\hat{A}_t,clip(r_t(\theta),1-\epsilon,1+\epsilon)\hat{A}_t) \right]\]

The policy output of the reinforcement learning has the dimension of 2, which ranges from -1 to 1. It gives the \(x,y\) displacement in every time step.
Observation dimension is \((128,128,3)\) what to describe the RGB data subscribed from the hand camera.

\subsection{Real2Sim and Sim2Real Domain Adaptation} 

    \begin{figure}
        \centering
        \includegraphics[trim=0 50 0 0, width=0.8\linewidth]{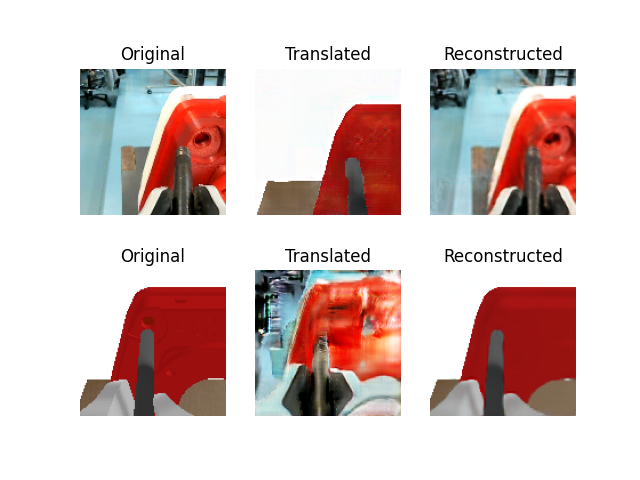}
        \caption{Sim and Real Transfer}
        \label{CycleGAN in Insertion Task}
    \end{figure}

A generative adversarial network (GAN) \cite{gan2014}\cite{cyclegan} has drawn wide attention in recent years, and there has shown wide applications of the GAN model in the industry, like image domain transfer or sim2real learning\cite{rlcyclegan}. In Fig \ref{CycleGAN in Insertion Task}, the original image from the real domain can be transferred into the simulation domain with the model \(G_{ab}\), and with \(G_{ba}\) it will be transferred back to the real domain. Training objective cycle GAN is to reduce the discrimination loss by updating the Discriminator while increase the discrimination loss with updating the Generator.


    
And in our solution, we do not only implement the sim2real in our training for better robustness, but also adapt the advantage of real2sim as well. Our robot aims to have the capability to transfer the real scenario to the simulation scenario, which narrows the gap between the realistic and simulation (Fig \ref{fig:domain_adaptation_and_mix}).

\subsection{Task Mixing, Randomization, and Vectorization}
In the simulation, multi-tasks have been designed for better and more robust learning performance. Paralleled training with multi-process is adopted in this approach, which significantly speeds up the training efficiency, from 15 fps to 200 fps (depending on the number of CPU cores). Details of each task can be found in Section \ref{task design}. Task randomization and mixing adopts the advantage of meta-learning \cite{schoettler2020metareinforcement}\cite{meta2017}, which helps the algorithm to learn more important and transferable features. 

\subsubsection{Domain Randomization}
Domain randomization \cite{tobin2017randomdomain} has been widely used in the real2sim transfer, seeing related work in Section \ref{literature_review} .

\subsubsection{Vectorization and Mixing- Efficient Paralleled Learning}
Vectored multiple environments and various tasks aim to increase the robustness of our algorithm. Multiple environments are running in every step of action prediction. Actions are vectored and passed through all \(n\) environments, and it will return \(n\) observations \cite{stable-baselines}. Multiple CPU will be used for environments for image rendering, which significantly increase the training efficiency.

\begin{figure}
    \centering
    \includegraphics[width=0.3\linewidth]{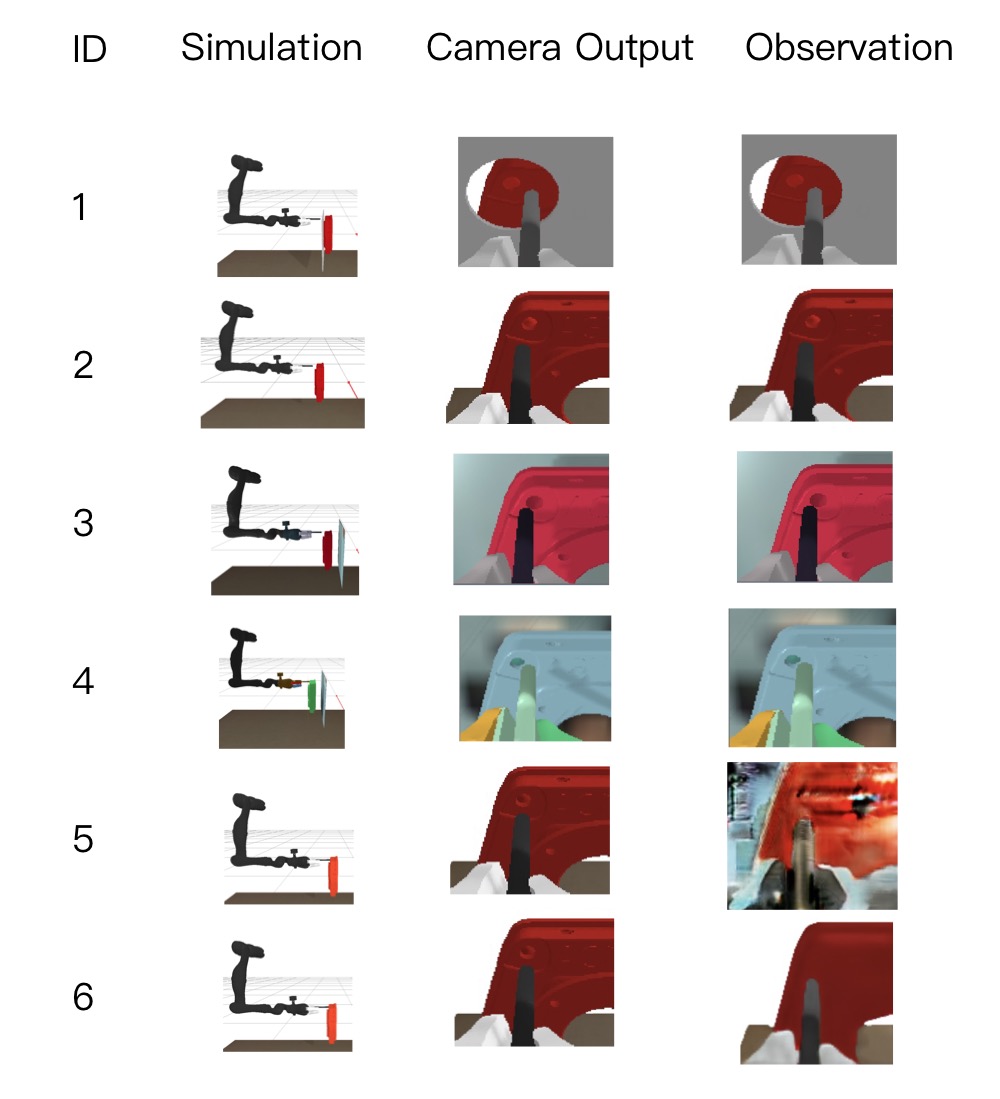}
    \caption{Simulation Task Designs}
    \label{fig:task_design}
\end{figure}

\section{Experiment}

\subsection{Experiment Overview}

Our experiments are mainly given out in following areas to show the result: test in the simulation, the performance improvement with vectored environments, mixing learning method, and the performance under real scenario.

\subsection{Task Design\label{task design}}
We have designed 6 tasks for training, as discribed in following list and Fig \ref{fig:task_design}.
%
    Task 1: It narrows the path search field using a searching allowed area. The initial position of the gripper is fixed. Generally, task 1 is a simplified task for the insertion. The disadvantage of this task is that CNN may not observe enough information, and it doesn't have a good capability to defeat noise, such as light and background variation.
    Task 2: In task 2,3,4 the front protect shell is invisible but gives a collision. The camera is able to capture the image of the whole scenario. In task 2, the insert behavior starts with a random position.
    Task 3: In this task, the gear box will change its color, following the idea of domain randomization \cite{tobin2017randomdomain}
    Task 4: It is a more difficult task than Task 3. In this task, the case color, bolt color, light, background color, texture will change randomly in a larger range.
    Task 5: Generative Model is used in this task. All the image input will be transferred to the domain of Realistic Images. The RL agent will access the "fake real" image in this task to learn actions. An important feature such as the hole might disappear in the generated model. This will be discussed in the Section \ref{import_feature_dis_discuss}.
    Task 6: Generative model is used to transfer the image from \(b\) to \(a\), and successively from \(a\) to \(b\). This task aims to help the agent to learn the \(a\) domain features given by the generative model, which is slightly different from the real simulation environment image (such as in Fig \ref{CycleGAN in Insertion Task}).
    Task 7: It generates the noise by GAN network, aiming to help RL learning under a uncertain environment.

\begin{figure}
    \centering
    \includegraphics[width=0.4\linewidth]{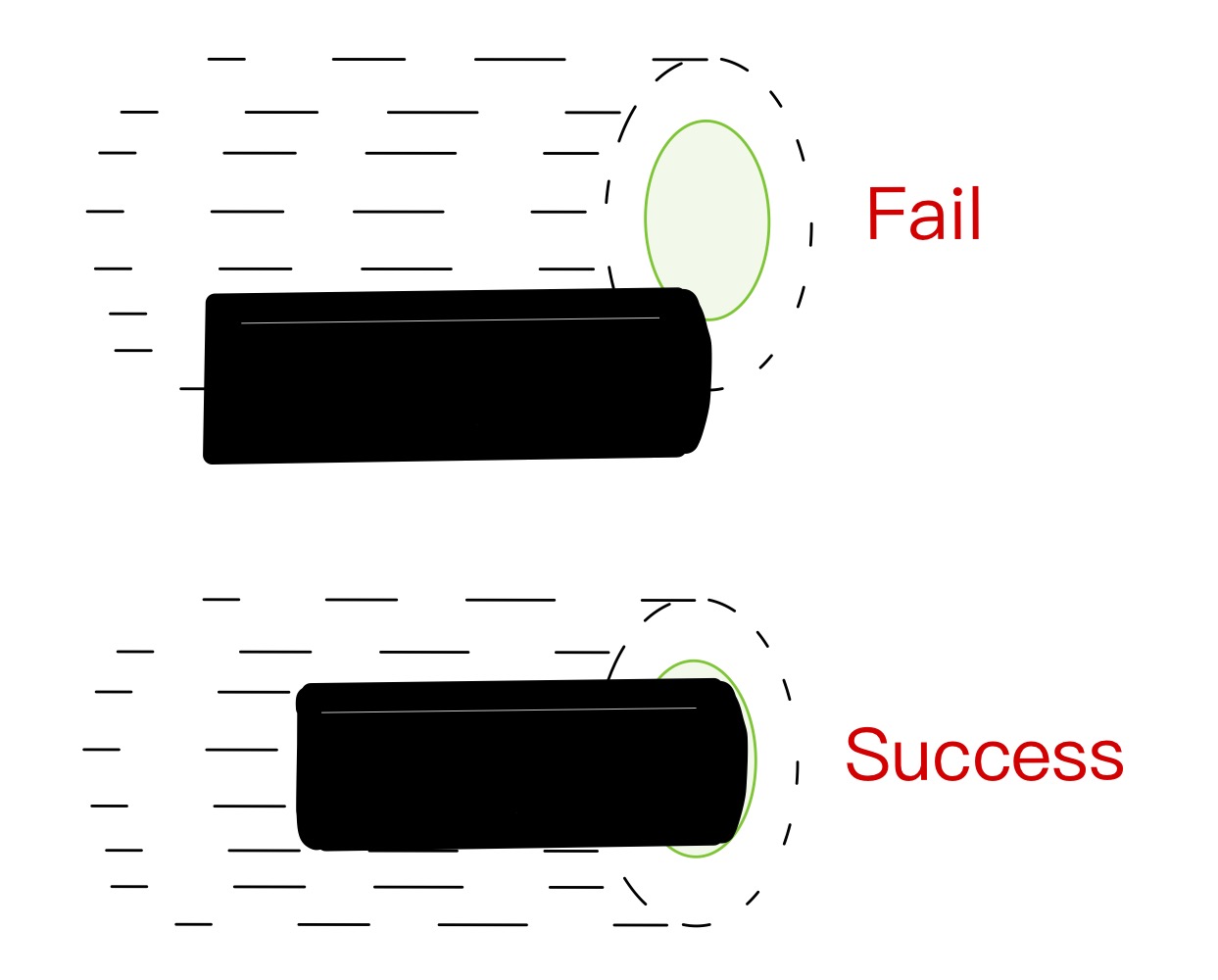}
    \caption{Terminal Condition}
    \label{Terminal Condition}
\end{figure}

\subsubsection{Terminal Condition}
Terminal signal will be triggered. If the bolt touches the insertion detector placed at the 0.5 cm depth of the hole, it will send the success signal to the simulation that the insertion is done.



\begin{figure}
    \centering
    \includegraphics[trim=100 0 100 0,width=0.7\linewidth]{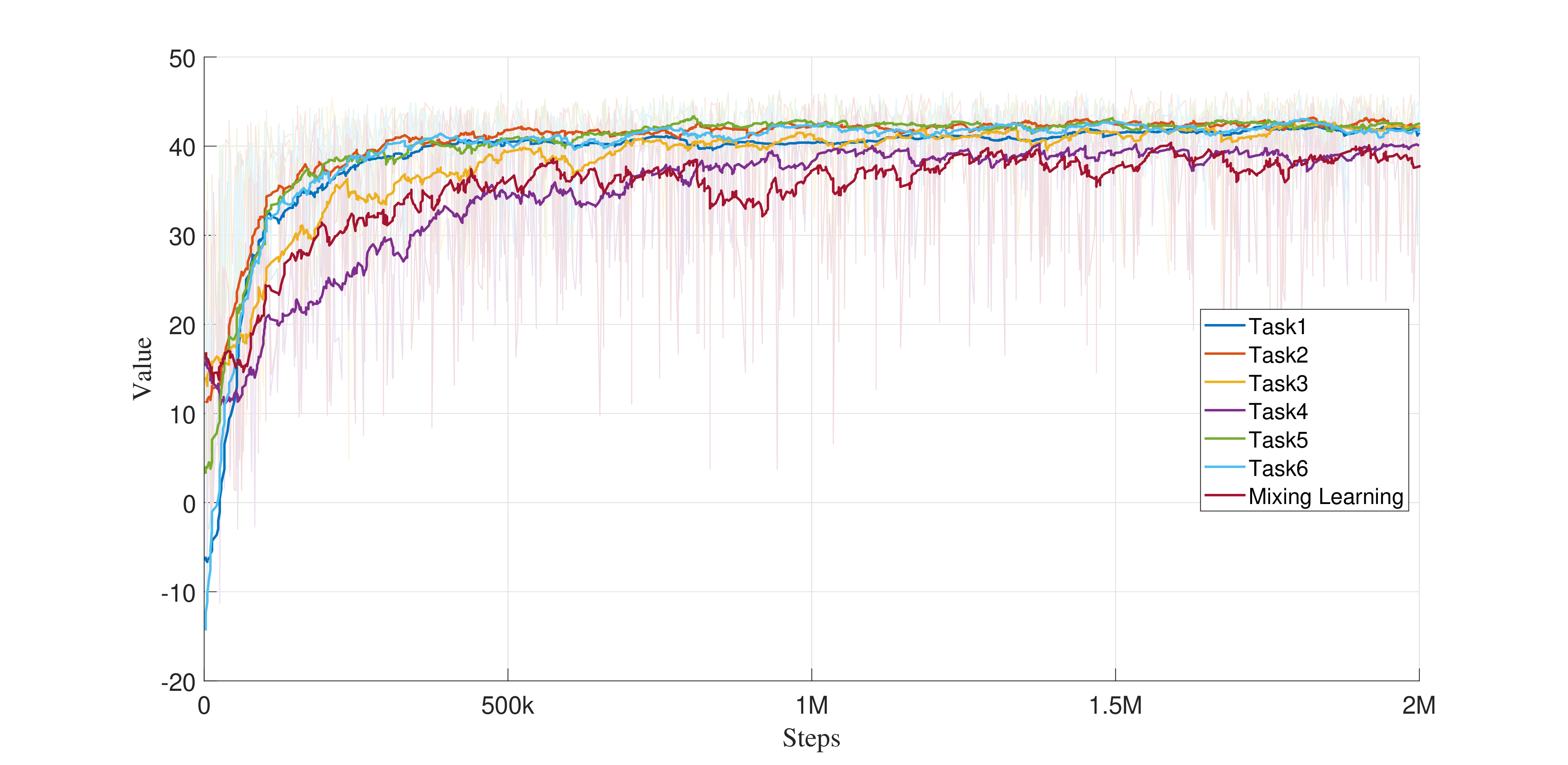}
    \caption{Success on different Tasks}
    \label{convergance}
\end{figure}

\subsection{Convergence of Reinforcement Learning on Tasks}
We test PPO algorithm under our different tasks, and the results are shown in the Fig \ref{convergance}.
It shows a good stability to solve all of these 6 task. And the task 7 is only for noise adding, and it will not be tested seperately. Mixing learning task introduced in section \ref{mixing learnin} is also tested. As the curve shows, task 4 shows a higher difficulty level, which costs more time to converge to the optimal policy.

\subsection{Mixing Learning\label{mixing learnin}}
\begin{figure}
    \centering
    \includegraphics[width=0.6\linewidth]{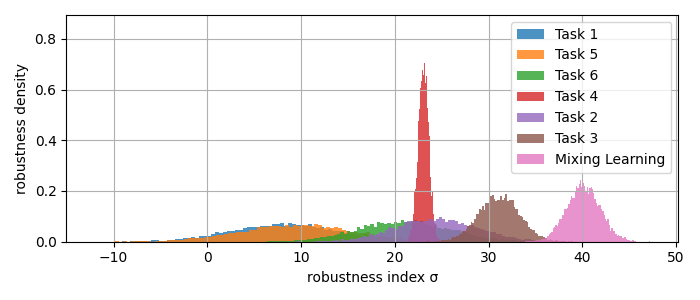}
    \caption{Robustness Test with Different Learning Methods}
    \label{robusttest}
\end{figure}

For testing the mixing learning for the performance improvement, we first give metrics to evaluate the robustness, noted as \(\sigma\). These metrics are relative to the robustness to solve all the tasks. A higher robustness index shows a better probability to solve wider tasks. It also illustrates the average return per episode per task.
Seeing in the Fig \ref{robusttest}, with mixing learning methods, the RL agent shows the highest robustness dealing with various tasks. Therefore, we adopt this model in real scenario test.

\[\sigma=\frac{\sum_{T=0}^{T_{max}}\sum_{e=0}^{e_{max}}\sum_{t=0}^{t_{max}}r_i }{T_{max}e_{max}}\]

\vspace{-0.5cm}

\subsection{Test in Real Scenario and Real2Sim}
\begin{figure}
    \centering
    \includegraphics[width=0.5\linewidth]{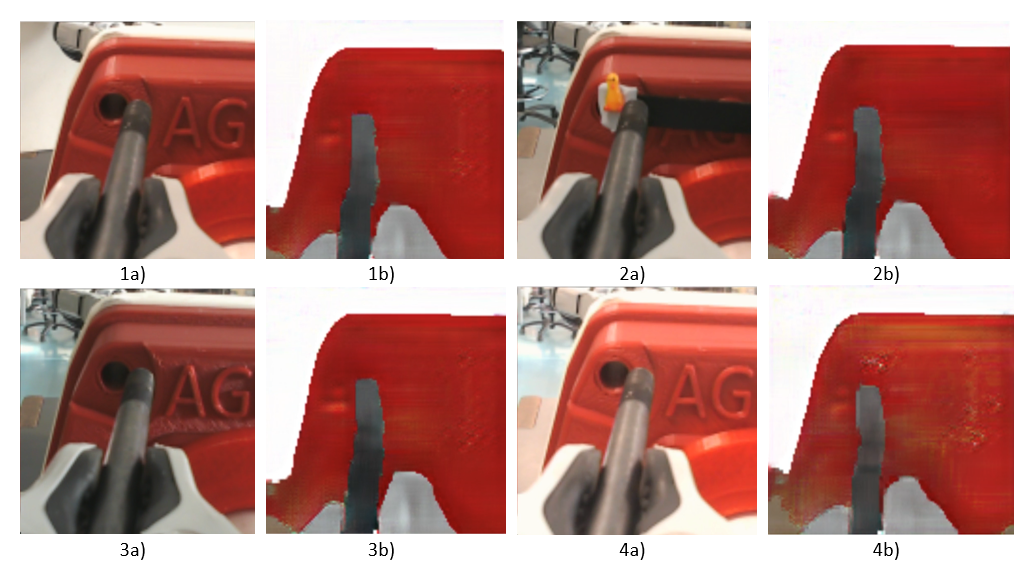}
    \caption{Real2Sim Adaptation and Noise Cancellation}
    \label{fig:real2sim and noise cancel}
\end{figure}

%
The work pipeline is introduced above in Fig \ref{fig:domain_adaptation_and_mix}. With this pipeline, we tested 13 times of insertion under a real scenario, and it shows a successful insertion with 12 times, with a success rate of 92.3\%. Fine-tuning of camera position is necessary for reproducing this result, since policy learning is a camera position-specific policy (see the discussion in Section \ref{CameraFixed} below).
Real2Sim is necessary for this pipeline. Policy learned from the simulation can totally lose its functionality in the real, since the real feature appears totally differently.
A good performance of real-time noisy background cancellation is observed. 
In this experiment, we intentionally give camera noise by background noise, front noise and light changing noise. As shown in the Fig \ref{fig:real2sim and noise cancel}, all the noise can be cancelled and show a good performance of Real2Sim transferring.

\subsection{Benchmark Comparison}

The comparison of our method with others is shown in Table \ref{Table}.
%
HLI (Human Labor Intensity) includes data labeling (scores 2 points) and human demonstration (scores 1 point). EC (Estimated Cost) gives this approach's cost (RGB Camera \$10, RGB-D Camera \$350, Torque/Force Sensors \$500). The lower MIS (Minimum Infrastructure Score) score requires less cost and setup complexity (with RGB camera = 1 point, T/F sensor =2 points, RGB-D camera = 2 points). MAS (Minimum Additional-information Score)  indicates to what degree the methods rely on hard to get information. The lower MAS score requires less hard to get information (with RGB image = 1 point, Torque/Force = 2 points, Target-End Effector Pose or Target-Pose Image = 3 points).  The final score of MIS or MAS is divided by assuming all subelements are included. In VNR, three factors are considered to measure the visual noise resistance of a method. The resistance to various background changes, environmental light changes, and camera color bias all score 1 point. Benchmark scores are given with our best knowledge of their methods.

According to the table result, our method's advantages lie in these aspects: First, our method performs well without any human intervention, while it is required of other methods, such as data labeling or human demonstrations, which are done manually and time-consuming. Second, We have successfully applied the sim2real in simulation learning-based method. In contrast, some of the others directly train robots in reality that security risks remain when training on real robots and the sampling efficiency is poor. Third, our method realizes insertion tasks with minimum infrastructure and additional information, only one RGB camera. In contrast, others are much more complex and expensive and bring huge financial burdens and hinder the industrial application. Forth, the VNR of our method scores full marks, which is much better than most other methods in resisting visual noise. In summary, our method achieves a relatively high success rate of insertion task while maintaining competitive position variability in a visually noisy environment and avoid the safety problems of real-machine training through the sim2real-based method with reduction of system complexity and cost.

\vspace{-0.5cm}
\begin{table*}[]\scriptsize
\setlength{\abovecaptionskip}{0.35cm}
\centering
\caption{\centering Benchmark Comparison. HLI represents human labor intensity. S2R represents whether the sim2real-based method is applied. EC represents an estimated cost. MIS and MAS represent minimal infrastructure scores and minimal additional information scores respectively. Acc means accuracy, PV denotes position variability and Prc. represents the precision of a method. VNR means visual noise resistance. 
The arrow up indicates that the score of the value of the indicator is positively correlated with the performance, representing the higher the better, vice versa.}
\resizebox{\textwidth}{!}{
\begin{tabular}{ c c c c c c c c c c c }
\hline
Methods                                                   & HLI $\downarrow$       & S2R          & EC $\downarrow$         & MIS $\downarrow$         & MAS $\downarrow$          & Acc $\uparrow$        & PV(mm) $\uparrow$     & Prc.(mm) $\downarrow$     & VNR $\uparrow$         & Sensors or Additional Info.         
\\ \hline
G. S. et al, \cite{marven1},2019                          & 2          & No           & 510\$         & 0.6          & 0.67          & 64\%$\sim$84\%  & 2           & 1             & 0/3          & Goal Pose image, RGB Cam., T/F \\ 
B. C. et al, \cite{camilo2020variable}, 2020                & 0          & \textbf{Yes} & 500\$         & 0.4          & 0.56          & 60\%$\sim$100\% & 5           & 1             & -            & T/F, Goal Pose      \\ 
M. V. et al, \cite{8794074}, 2019                            & 3          & No           & 510\$         & 0.6          & 0.33          & 77\%$\sim$97\%  & \textbf{40} & 0.5  & 2/3          & RGB Cam, T/F                   \\ 
J. X. et al, \cite{8454796}, 2019                                & 0          & No           & 500\$         & 0.4          & 0.22          & \textbf{100\%}  & 0           & 0.02 & -            & T/F                                \\ 
G. S. et al, \cite{schoettler2020metareinforcement}, 2020 & 2          & \textbf{Yes} & 850\$         & 0.8          & 0.67          & \textbf{100\%}  & 6           & 0.1           & -            & RGB-D Cam., T/F, Goal Pose   \\ 
T. I. et al, \cite{8}, 2017                                   & 0          & No           & 500\$         & 0.4          & 0.22          & \textbf{100\%}  & 3           & \textbf{0.01} & -            & T/F                                \\ 
X. C. et al, \cite{11}, 2020                                   & 2          & \textbf{Yes} & -           & -            & -             & 82\%            & -           & -             & 2/3          & Pick and Pouring tasks                    \\ 
A. H. et al, \cite{10}, 2020                              & 2          & \textbf{Yes} & \textbf{10\$} & \textbf{0.2} & \textbf{0.11} & 73\%$\sim$100\% & -           & 30            & \textbf{3/3} & \textbf{RGB Cam.}                       \\ 
\textbf{Ours}                                             & \textbf{0} & \textbf{Yes} & \textbf{10\$} & \textbf{0.2} & \textbf{0.11} & \textbf{92.3\%} & \textbf{30} & \textbf{1}    & \textbf{3/3} & \textbf{RGB Cam.}                       \\ \hline
\end{tabular}
\label{Table}}
\end{table*}

\subsection{Discussion and Future Work}

We discuss following issues and future works, Camera Position Fixed Policy\label{CameraFixed}, Important Feature Vanishing\label{import_feature_dis_discuss}, Force Detection and Fusion, Dedicated Observation Network Design, Higher Precise and Stereo Camera, Robustness.
1. The current observation network is based on vanilla CNN, it does not have the capability to deal with time-based problems. It is also not deep enough and the architecture is not well-tuned. It is also important to note that even though we adapt domain randomization in our tasks design, the camera position is set almost fixed (variance is quite small). A strong precondition is made that the relative position of the camera and gripper is assumed given. This precondition may cause strict effort given upon the tuning position and orientation of the camera to match the setup in the simulation. The policy will show decrease performance with the changing of the camera position or orientation. And this problem will also be addressed in our future work.
2. During the Cycle GAN training and inference, we found the important features can be lost during the domain transferring, which was also indicated by \cite{rlcyclegan}. Even though our approach shows it is capable to finish the insertion without those features, we believe in future work it is better to improve the visual network and GAN learning techniques.
3. Force detection and learning are not integrated into our algorithm, but which is important during the insertion and protects the hole quality. In future work, we are aiming to design a force fusing network to fuse the force information into reinforcement learning. 
4. Currently, the version of the observation network inherited from the advantage of nature CNN, a 3 layers vanilla network. The performance can be potentially improved if a deeper network is implemented.
5. Currently insertion precision in the real scenario is about 1mm clearance. We are also considering using a stereo camera to provide richer image information, aiming to mimic human insertion behavior. An intelligent agent with a stereo camera is able to learn the representation of distance, and that could potentially improve the insertion precision and robustness.
6. The robustness of this approach can be potentially further improved in future work.

\section{Conclusion}
We have proposed a practical approach to solve the real insertion task under a noisy environment, with only RGB camera information. We provide the novel transfer learning method Real2Sim, for fast and easier policy adaptation.
It shows a decent performance with 1mm clearance with fine-tuning of camera position. In future work, we will work on using stereo cameras and give a higher camera invariable robustness and precision.

\section{Acknowledgement}
This research is supported by the Agency for Science, Technology and Research (A*STAR), Singapore, under its AME Programmatic Funding Scheme (Project \#A18A2 \\
b0046). The computational work for this article was partially performed on resources of the National Supercomputing Centre, Singapore (https://www.nscc.sg). Thanks for the technical help from Jing Wei, Puang En Yen, Shanxiang Fang.

%
%

\end{document}